\documentclass[12pt]{article}
\usepackage{graphics}

\begin{document}

\title{\ \\ \LARGE\bf A New Monte Carlo Based Algorithm for the Gaussian Process Classification
Problem }

\author{Amir F. Atiya\\
Department of Computer Engineering\\
Cairo University\\
Cairo, Egypt\\
amir@alumni.caltech.edu\\
\\
\and
Hatem A. Fayed\\
Department of Engineering Mathematics and Physics\\
Cairo University\\
Cairo, Egypt\\
h\_fayed@eng.cu.edu.eg \\
\\
\and
Ahmed H. Abdel-Gawad\\
Department of Electrical Engineering\\
Purdue University\\
ah.hamdy@gmail.com\\
}


\maketitle

\begin{abstract}
Gaussian process is a very promising novel technology that has been
applied to both the regression problem and the classification problem.
While for the regression problem it yields simple exact solutions,
this is not the case for the classification problem, because 
we encounter intractable integrals. In this paper 
we develop a new derivation that transforms the
problem into that of evaluating the ratio
of multivariate Gaussian orthant integrals.
Moreover, we develop a new Monte Carlo procedure
that evaluates these integrals. It
is based on some aspects of bootstrap sampling and acceptance-rejection.
The proposed approach has beneficial properties compared
to the existing Markov Chain Monte Carlo approach,
such as simplicity, 
reliability,
and speed.
\end{abstract}

\section{Introduction}

Gaussian process classification (GPC) is a promising
Bayesian approach to the classification problem.
It is based on defining a so-called latent variable
for every pattern, and setting the prior based 
on proximity relations among the patterns.
Based on the prior and through a series of integrals,
the posterior of a test pattern will determine
its classification (see the reviews of Rasmussen and Williams \cite{RASMUSSEN05}, 
Nickisch and Rasmussen \cite{NICKISCH08},
and Seeger \cite{SEEGER04}).
Unfortunately, this resulting 
multi-integral formula is intractable,
and can only be solved through some approximations or using lengthy
algorithms. In spite of its superior performance (as has been
pointed out in several application studies, such as by Altun, Hofmann, and Smola \cite{ALTUN04},
Jenssen et al \cite{JENSSEN07}, and Bazi and Melgani \cite{BAZI10}), this computational
issue hampers its wide applicability to large data sets. In this
paper, we propose a new simplified, but exact, formula
for the binary GPC problem. The proposed method is based on applying some substitutions
and transformations that convert the problem into that of evaluating
a ratio of two orthant integrals of a multivariate Gaussian density. 
Moreover, we develop a new Monte-Carlo-type algorithm to evaluate these
orthant integrals. 
The proposed algorithm is based on some aspects of bootstrap sampling
and acceptance-rejection. 

The approaches in the literature for the GPC problem
can be categorized into solutions based on analytical approximation
of the integrals and solutions based on Monte Carlo sampling (see the extensive review
of Nickisch and Rasmussen \cite{NICKISCH08}). Among
the well-known proposed methods from the first category are the Laplace's
approximation (Williams and Barber \cite{WILLIAMS98}) and the expectation
propagation (Minka \cite{MINKA01}). Also, some other efficient approximation-based
methods include the work of Csat\'o et al \cite{CSATO00}, Opper and Winther \cite{OPPER00}), Gibbs
and MacKay \cite{GIBBS00}, Rifkin and Klautau \cite{RIFKIN04}, Jaakkola and Haussler \cite{JAAKOLA99}, 
and Kim, and Ghahramani \cite{KIM06}. Work on the second category has been more scarce. Almost all of
the approaches are based on the Markov Chain Monte Carlo (MCMC) concept. 
Neal's so-called 
{\it Annealed Importance Sampling (AIS)} \cite{NEAL98} 
uses an approximate posterior, rather than the prior, as a starting point
for evaluating the marginal likelihood (to be described shortly).
The {\it Hybrid Monte Carlo (HMC)}  (Neal \cite{NEAL99}) is based on the concept of ``importance density".
Murray, Adams, and Mackay \cite{MURRAY10} proposed the Elliptical Slice Sampler (ESS). It 
is based on sampling over an elliptical slice to efficiently 
obtain the step size.
Titsias, Lawrence and Rattrpay \cite{TITSIAS09} introduced a novel MCMC approach by making
use of a low dimensional vector of control variables (see also Titsias and Lawrence \cite{TITSIAS10}).
Vehtari, S\"arkk\"a, and Lampinen \cite{VEHTARI00}  proposed a novel way to
choose effective starting values for the MCMC based on early stopping.
Barber and Williams \cite{BARBER97} proposed a hybrid method, that is
uses partly an approximation and partly the MCMC procedure. 

Nickisch and Rasmussen \cite{NICKISCH08}) provided a comprehensive review and comparison
between the different GPC approximations and the MCMC approaches. 
They came to the conclusion that the MCMC-type approaches are superior in performance,
but of course computationally more extensive. This is because they can obtain
the exact solution of the integral formula, provided the size of the run is large enough.

Kuss and Rasmussen \cite{KUSS05} compared between a number of GPC approximations.
Rather than improving the approximation ability or the computation speed,
some researchers considered short-cuts to the the GPC model itself
to achieve better efficiency or performance, such as sparse GPC models (see Csat\'o and Opper \cite{CSATO02}
Vanhatalo and Vehtari \cite{VANHATALO10}, and Titsias and Lawrence \cite{TITSIAS09b}), and low-dimensional manifold embedding
(see Ursatun and Darrell \cite{URSATUN07}).

A majority of the work in the literature and the above reviewed methods consider the binary classification case.
Nevertheless, some researchers extended The GPC problem
to the multi-class case (for example Girolami and Rogers \cite{GIROLAMI06}, Hern\'andez-Lobato et al \cite{HERNANDEZ}
and Seeger and Jordan \cite{SEEGER04b}).

The method proposed in this paper fits more into the second category described above (i.e. exact Monte Carlo-based), but it is not based on the MCMC concept.
It is guaranteed to converge as close as possible to the exact solution of the multi-integral formula,
provided we use a large enough sample of generated points. 
The advantages of the proposed
algorithm is that it does not require any parameter-tuning (other than specifying the number
of Monte Carlo generated points), is consistent, and reliable.
(In short it works all the time, we tested hundreds of problems, some as high as 2000 dimensional problems.)
It also compares favorably in terms of speed and accuracy to the other MCMC approaches, especially
for the evaluation of the marginal likelihood. 
The marginal likelihood is an expression for the likelihood of the data given the 
parameters. The hyperparameters are typically tuned by optimizing the marginal likelihood function.
Because of repeated evaluations of the likelihood function, this step
is the most time-consuming part, and  the speed-up provided by the
proposed algorithm will lead to a significant computational benefit.
A beneficial aspect of the proposed integral formulation 
is that it gives many insights into the
different influencing factors. For example, one can obtain
the limiting behavior of the covariance matrix parameters,
and therefore understand the classification behavior when moving their values
in certain directions. Also, the other advantage of the integral
formulation is that it is given in terms of multivariate
Gaussian orthant integrals. These are well-researched integrals,
and several approximations exist in the literature. So, this could possibly
 open the way for new competitive approximations to the GPC problem.
 
 The paper is organized as follows. Next section we present an overview and definition
 of the GPC problem. In Section 3 we propose the new formulation that is obtained by
simplifying the multi-integral formula. Section 4 presents an overview
of the multivariate Gaussian integral that has to be evaluated. 
In Section 5 we propose the new Monte Carlo model for evaluating
the integral. Section 6 provides the experimental study to assess the new model,
and Section 7 is the conclusion.

\section{The Gaussian Process Classification Problem}

Gaussian process classifiers (GPC) are based on defining a ``latent
state" $f_i$ for every training pattern. It is a central variable in the formulation
which measures some sort of degree of membership to one of the classes.
Let $y_i$ denote the class membership of training pattern $i$,
where $y_i=1$ denotes Class 1 and $y_i=-1$ denotes Class 2.
The latent variable $f_i$, whose
range is from $-\infty$ to $\infty$, is mapped into class posterior probability through a monotone 
squashing function $\sigma$ that has a range of $(0,1)$, as follows. 
\begin{equation}
J=P(y_{i}=1|f_{i})=\sigma(f_{i})\label{eq:sigma}\end{equation}
There are two typical forms for $\sigma$ in the GPC literature: the logit (or logistic function)
and the probit (or cumulative Gaussian integral). As argued by 
Nickisch and Rasmussen \cite{NICKISCH08}, both choices are effectively quite similar.
In this work we
consider only the probit function.

In what is next we will follow closely
the terminology of Rasmussen and Williams \cite{RASMUSSEN05}.
Let us arrange the latent variables and the class memberships in one vector each:
${\bf f}=(f_1,\dots,f_N)^T$, and ${\bf y}=(y_1,\dots,y_N)^T$.
Note that each index of the afforementioned vectors pertains to a specific training pattern,
and $N$ is the size of the training set.  
Let ${\bf x}_i$ be the feature vector of training pattern $i$.
Moreover, let us arrange all training vectors ${\bf x}_{i}$ as rows in 
a matrix $X$.  Let ${\bf x}_{*}$ denote the feature vector of the test pattern,
whose class needs to be evaluated. Let its latent state be $f_{*}$.

 The latent state vector $\bf f$ obeys an a priori
density that is assumed to be a multivariate Gaussian (therefore the
name Gaussian processes). From an a priori point of view, patterns that are close (in the features space)
are more likely to belong to the same class. So this prior density
is selected to reflect that property. Patterns with nearby feature vectors have
highly correlated latent variables $f_i$, and as the patterns become more distant the correlation
decays.  The a priori density can be written as

\begin{equation}
p({\bf f}|X)={\cal N}({\bf f};0,\Sigma)\label{eq_P_F_G_X}\end{equation}
where ${\cal N}({\bf f};{\bf \mu},\Sigma)$ denotes a Gaussian density of variable ${\bf f}$
having mean vector ${\bf \mu}$ and covariance matrix $\Sigma$. The covariance
matrix has elements that are a function of the distance between two
feature vectors $\| {\bf x}_{i}-{\bf x}_{j}\|^{2}$ and is so designed to achieve
this aforementioned correlation behavior (see Rasmussen and Williams \cite{RASMUSSEN05} for a detailed
discussion and examples of covariance functions). A particularly prevalent choice
of the covariance matrix is the so called ``RBF" covariance matrix,
given by:
\begin{equation}\label{RBF}
\Sigma_{ij}=\beta e^{-{\| {\bf x}_{i}-{\bf x}_{j}\|^{2}}\over{\alpha^2}}
\end{equation}
The $\alpha$ and $\beta$ parameters (called respectively the length scale
and the latent function scale) are very influential 
in the performance of the classifier, and tuning them
has to be done with care (see Sundarajan and Keerthi \cite{SUNDARAJAN01}).
More will be said later on them.

A test pattern's latent variable $f_*$ will have similar correlation structure
as the training patterns. Consider the augmented training latent state vector and test point latent
state. It is given as Gaussian, as follows:

\begin{equation} \label{F-FSTAR}
\left[\begin{array}{c}
{\bf f}\\
f_{*}\end{array}\right]\sim {\cal N}\left(\left[\begin{array}{c}
{\bf f}\\
f_{*}\end{array}\right];0,\left[\begin{array}{cc}
\Sigma & \Sigma_{X {\bf x}_{*}}\\
\Sigma_{X {\bf x}_{*}}^T & \Sigma_{{\bf x}_{*}{\bf x}_{*}}\end{array}\right]\right)\label{eq_f_fstar}\end{equation}
where ${\Sigma}_{X {\bf x}_{*}}$ is the covariance 
between the training latent variables and the test latent variable (it is a vector), and
$\Sigma_{{\bf x}_{*}{\bf x}_{*}}$ is the variance of $f_{*}$. 
A key to estimating the class membership of the test point is to evaluate
the probability density of its latent state $f_{*}$, conditional
on all the information that is available from the training set: 

\begin{equation}
p(f_{*}|X,{\bf y},{\bf x}_{*})=\int p(f_{*}|X,{\bf x}_{*},{\bf f})p({\bf f}|X,{\bf y})d{\bf f}\label{eq_Rasmussen1}\end{equation}
where $p({\bf f}|X,{\bf y})=\frac{p({\bf y}|{\bf f})p({\bf f}|X)}{p({\bf y}|X)}$ is the posterior of
the latent variables.

Then, we compute the probability of Class 1 averaged over the
conditional density of $f_{*}$:

\begin{equation} \label{J-STAR}
J_{*}\equiv p(y_{*}=+1|X,{\bf y},{\bf x}_{*})=\int\sigma(f_{*})p(f_{*}|X,{\bf y},{\bf x}_{*})df_{*}\label{eq_P_Yst_G_X_Y_Xst}\end{equation}
where we used the fact that $\sigma(f_*)$  signifies the conditional given in Eq. (\ref{eq:sigma}). We get

\begin{equation}
p(f_{*}|X,{\bf y},{\bf x}_{*})={ {\int p(f_{*}|X,{\bf x}_{*},{\bf f})p({\bf y}|{\bf f})p({\bf f}|X)d{\bf f}}\over{p({\bf y}|X)}}\label{eq_P_Fst_G_X_Y_Xst}\end{equation}
where

\begin{equation}
p({\bf y}|{\bf f})=\prod_{i=1}^{N}p(y_{i}|f_{i})=\prod_{i=1}^{N}\sigma(y_{i}f_{i})=\prod_{i=1}^{N}\int\limits _{-\infty}^{y_{i}f_{i}}\frac{e^{-\frac{x^{2}}{2}}}{\sqrt{2\pi}}dx,\label{eq_P_Y_G_F}\end{equation}
where we used the fact that $\sigma$ is the probit function
(integral of the Gaussian function). Note that $P(y_i=-1|f_i)=1-P(y_i=1|f_i)=1-\sigma({f_i})=\sigma(-f_i)=\sigma(y_i f_i)$
because of the point symmetry of $\sigma$.
Also,

\begin{equation}
p(f_{*}|X,{\bf x}_{*},{\bf f})={\cal N}(f_{*};{\bf a}^{T}{\bf f},\sigma_{*}^{2})\label{eq_P_Fst_G_X_Xst_F}\end{equation}
where
\begin{equation}\label{a-DEF}
{\bf a}=\Sigma^{-1}\Sigma_{X {\bf x}_{*}}
\end{equation}

\begin{equation}\label{sigstar}
\sigma_{*}^{2}=\Sigma_{{\bf x}_{*}{\bf x}_{*}}-\Sigma_{X {\bf x}_{*}}^T\Sigma^{-1}\Sigma_{X {\bf x}_{*}}
\end{equation}
We utilized formulas expressing the conditional of a multidimensional Gaussian distribution,
applied to  Eq. (\ref{eq_f_fstar}).

All past formulas follow from straightforward probability manipulations, and they are
 described clearly in Rasmussen and Williams \cite{RASMUSSEN05}, p. 16 Eq. 2.19. 
Equation (\ref{J-STAR}), representing the posterior probability corresponding to
the test pattern ${\bf x}_*$, is the main quantity needed to classify the pattern.
Thus, $J_*\ge 0.5$ means that the pattern should be classified as Class 1,
and otherwise it should be classified as Class 2. 

Another important quantity
that is needed is the so-called marginal likelihood $L$, defined as

\begin{equation}\label{MARG-LIK}
L=p({\bf y}|X)
\end{equation}
It is an important quantity for the purpose of tuning the two hyperparameters ($\alpha$ and $\beta$).
By maximizing the marginal likelihood, we arrive
at hyperparameters that are most consistent with the observed data.
As such, any method should also be able to efficiently evaluate
the marginal likelihood. See Rasmussen and Williams \cite{RASMUSSEN05} for more information
about the marginal likelihood.

\section{The Proposed Simplification of the Multi-Integral}
\subsection{Variable Transformation}
Evaluating Equations (\ref{eq_P_Yst_G_X_Y_Xst}), (\ref{eq_P_Fst_G_X_Y_Xst})
analytically is very hard to accomplish. The difficulty arises also
when attempting to evaluate them numerically because of the high dimensionality
of the integrals (for example for a problem with a training set of
size 1000 we are dealing with more than a thousand-fold integral). Also, attempting
a standard Monte Carlo evaluation leads to some practical problems,
among them is the fact that $\prod_{i=1}^{N}\sigma(y_{i}f_{i})$ turns
out to be usually a very small number (with a negative exponent with
a very large magnitude). So to summarize, we are dealing with a very hard
problem if an exact solution is to be sought. Here we develop a procedure that transforms
the problem into the more approachable form of integrals of multivariate Gaussian functions. 
Specifically, we perform some substitutions and transformations
of variables that will convert the problem into evaluating orthant
integrals of some multivariate Gaussian density. By orthant integral
we mean an integral of a zero-mean multivariate Gaussian function
over some quadrant, e.g. over ${\bf x}\ge0$. The detailed steps
are given below.

Substituting  (\ref{eq_P_Y_G_F}), (\ref{eq_P_F_G_X}), (\ref{eq_P_Fst_G_X_Xst_F})
and (\ref{eq_P_Fst_G_X_Y_Xst}) into  (\ref{eq_P_Yst_G_X_Y_Xst}), we obtain:

\begin{eqnarray}\nonumber
J_{*}&=&{1\over{p({\bf y}|X)}}\int\limits _{f_{*}=-\infty}^{\infty} \Biggl[ \int\limits _{u=-\infty}^{f_{*}}\frac{e^{-\frac{u^{2}}{2}}}{\sqrt{2\pi}}du\Biggr] \;\int_{f}\Biggl[\prod\limits _{i=1}^{N}\int\limits _{z_{i}=-\infty}^{y_{i}f_{i}}\frac{e^{-\frac{z_{i}^{2}}{2}}}{\sqrt{2\pi}}dz_{i}\Biggr]\\
&\ & {\cal N}({\bf f};0,\Sigma)\; {\cal N}(f_{*};{\bf a}^{T}{\bf f},\sigma_{*}^{2})df_{1}df_{2}..df_{N}df_{*}
\label{eq_Jst1}\end{eqnarray}
Rearranging, we get
\begin{equation}
J_{*}=\frac{
\int\limits _{f_{*}=-\infty}^{\infty}\int\limits _{u=-\infty}^{f_{*}}\int\limits _{f_{1}=-\infty}^{\infty}...\int\limits _{f_{N}=-\infty}^{\infty}\int\limits _{z_{1}=-\infty}^{y_{1}f_{1}}...\int\limits _{z_{N}=-\infty}^{y_{N}f_{N}}
 e^{-\frac{W}{2}} dz_{1}...dz_{N}df_{1}...df_{N}dudf_{*}
}{(2\pi)^{N+1}\sigma_{*}|\Sigma|^{\frac{1}{2}}p({\bf y}|X)}
\label{eq_Jst2}
\end{equation}
where


\begin{equation}
W=u^{2}+\sum_{i=1}^{N}z_{i}^{2}+{\bf f}^{T}\Sigma^{-1}{\bf f}+\frac{(f_{*}-{\bf a}^{T}{\bf f})^{2}}{\sigma_{*}^{2}}\label{eq_WSimple}\end{equation}
Rewriting $W$ in matrix form:

\begin{eqnarray*}
W&=&\left[\begin{array}{cccc}
u & z_{1} & \cdots & z_{N}\end{array}\right]\left[\begin{array}{c}
u\\
z_{1}\\
\vdots\\
z_{N}\end{array}\right]+{\bf f}^{T}\Sigma^{-1}{\bf f}+\\
& &\left[\begin{array}{cccc}
f_{*} & f_{1} & \cdots & f_{N}\end{array}\right]B\left[\begin{array}{c}
f_{*}\\
f_{1}\\
\vdots\\
f_{N}\end{array}\right]\label{eq_W}\end{eqnarray*}
where $B=\frac{1}{\sigma_{*}^{2}}\left[\begin{array}{c}
1\\
-{\bf a}\end{array}\right]\left[\begin{array}{cc}
1 & -{\bf a}^{T}\end{array}\right]$. The problem with this integral ($J_{*}$) is that some of its variables
($f_{*}$, $f_{1}$, ..., $f_{N}$ ) occur in the limits of the integrals.
A transformation can fix this problem by using the substitution (see the preliminary work of \cite{GAWAD08}):

\begin{equation}
{\bf v}\equiv\left[\begin{array}{c}
v_{1}\\
v_{2}\\
\vdots\\
v_{2N+2}\end{array}\right]=\left[\begin{array}{cccccccc}
-1 & 0 & 0 & 1 & 0 & 0 & 0\\
0 &  -1 & 0 & 0 & y_{1} & 0 & 0\\
0 &  \ddots & 0 & 0 & 0 & \ddots & 0\\
0 & 0 & -1 & 0 & 0 & 0 & y_{N}\\
0 & 0 & 0 & 1 & 0 & 0 & 0\\
0 & 0 & 0 & 0 & 1 & 0 & 0\\
0 & 0 & 0 & 0 & 0 & \ddots & 0\\
0 & 0 & 0 & 0 & 0 & 0 & 1\end{array}\right]\left[\begin{array}{c}
u\\
{\bf z}\\
f_{*}\\
{\bf f}\end{array}\right]\label{eq_Substs}\end{equation}
We get


\begin{equation}
J_{*}=\frac{
\int\limits _{v_{1}=0}^{\infty}\int\limits _{v_{2}=0}^{\infty}...\int\limits _{v_{N+1}=0}^{\infty}\int\limits _{v_{N+2}=-\infty}^{\infty}...\int\limits_{v_{2N+2}=-\infty}^{\infty}
 e^{-\frac{{\bf v}^{T}D{\bf v}}{2}}dv_{1}\cdots dv_{2N+2}
}{(2\pi)^{N+1}\sigma_{*}|\Sigma|^{\frac{1}{2}}p({\bf y}|X)}
\label{eq_Jst_Wdsh}
\end{equation}
where 

\begin{equation}\label{DMATRIX}
D=\left[\begin{array}{cccc}
1 & 0 & -1 & 0\\
0 & I & 0 & -C'\\
-1 & 0 & 1+\frac{1}{\sigma_{*}^{2}} & -\frac{{\bf a}^{T}}{\sigma_{*}^{2}}\\
0 & -C' & -\frac{{\bf a}}{\sigma_{*}^{2}} & I+\Sigma^{-1}+\frac{{\bf a}{\bf a}^{T}}{\sigma_{*}^{2}}\end{array}\right]\label{eq_D}\end{equation}
where $C'=\left[\begin{array}{cccc}
y_{1}\\
 & y_{2}\\
 &  & \ddots\\
 &  &  & y_{N}\end{array}\right]$ and $I$ is the identity matrix (in both cases in the formula it
is $N\times N$). The integration can  then be put in the form:

\begin{equation}
J_{*}=\frac{1}{|D|^{\frac{1}{2}}|\Sigma|^{\frac{1}{2}}p({\bf y}|X)\sigma_{*}}\int {\cal N}\left({\bf v};0,D^{-1}\right)d{\bf v}\label{eq_Jst_N_VOD}\end{equation}
This integral above is called orthant normal integral where the orthant
is defined over $\left[\begin{array}{c}
v_{1}\\
\vdots\\
v_{N+1}\end{array}\right]\ge0$ and $-\infty<\left[\begin{array}{c}
v_{N+2}\\
\vdots\\
v_{2N+2}\end{array}\right]<\infty$. Let us denote this orthant by {\it orth}. Consider now the term $p({\bf y}|X)$.
Integrating (\ref{eq_P_Fst_G_X_Y_Xst}) w.r.t. $f_{*}$ from $-\infty$
to $\infty$, and using the fact that $\int p(f_{*}|X,{\bf y},x_{*})df_{*}=1$
we get

\begin{equation}\label{PYX}
\begin{array}{cl}
p({\bf y}|X) & =\int\limits _{-\infty}^{\infty}\int_{{\bf f}}p(f_{*}|X,{\bf x}_{*},{\bf f})p({\bf y}|{\bf f})p({\bf f}|X)d{\bf f}\: df_{*}\\
 & =\int\limits _{-\infty}^{\infty}\int\limits _{-\infty}^{\infty}\frac{e^{-\frac{x^{2}}{2}}}{\sqrt{2\pi}}dx\int_{{\bf f}}p(f_{*}|X,{\bf x}_{*},{\bf f})p({\bf y}|{\bf f})\\
& p({\bf f}|X)d{\bf f}\: df_{*}\end{array}\label{eq_Integral_P_Y_G_X}\end{equation}

The  integral $\int\limits _{-\infty}^{\infty}\frac{e^{-\frac{x^{2}}{2}}}{\sqrt{2\pi}}dx$
in the previous equation is inserted on purpose. It equals 1 so it will not alter the formula. The
above integral will be the same as (\ref{eq_Jst_N_VOD}) except that
the limits will be over the orthant defined by $\left[\begin{array}{c}
v_{2}\\
\vdots\\
v_{N+1}\end{array}\right]\ge0$ and $-\infty<\left[\begin{array}{c}
v_{1}\\
v_{N+2}\\
\vdots\\
v_{2N+2}\end{array}\right]<\infty$. The reason is that the above analysis that led to (\ref{eq_Jst_N_VOD})
will apply here except that one of the integral's limits is different
(the Gaussian integral with variable $x$ that is inserted in (\ref{eq_Integral_P_Y_G_X}),
here the upper limit is $\infty$ instead of $f_{*}$). Let us denote
this orthant by {\it orth+}. The expression for the posterior will then
be given as.

\begin{equation}
J_{*}=p(y_*=1|X,{\bf y},{\bf x}_{*})=\frac{\int_{orth}{\cal N}\left({\bf v};0,D^{-1}\right)d{\bf v}}{\int_{orth+}{\cal N}\left({\bf v};0,D^{-1}\right)d{\bf v}}=\frac{I_{1}}{I_{2}}\label{eq_I1_I2}\end{equation}

\subsection{Further Reduction: }

The limits for the portion $(v_{N+2},\dots,v_{2N+2})^T$ in Expression (\ref{eq_I1_I2})
are from $-\infty$ to $\infty$, so these variables can be integrated
out. This means that the expression can further be reduced from a $2N+2$
dimensional integral to an $N+1$ dimensional integral. The detailed steps
of this reduction are given below.

Let ${\bf v}=\left[\begin{array}{c}
v_{1}\\
\vdots\\
v_{N+1}\end{array}\right]$ and ${\bf v}'=\left[\begin{array}{c}
v_{N+2}\\
\vdots\\
v_{2N+2}\end{array}\right]$. We can write:

\begin{equation}
I_{1}=k_{3}\int\limits _{{\bf v}\ge0}\int\limits _{-\infty\le {\bf v}'\le\infty}\frac{e^{-\frac{1}{2}\left({\bf v}^{T}A_{11}{\bf v}-2{\bf v}^{T}A_{12}{\bf v}'+{\bf v}'^{T}A_{22}{\bf v}'\right)}}{(2\pi)^{\frac{N+1}{2}}|A_{22}|^{-\frac{1}{2}}} d{\bf v}' d{\bf v}\label{eq_Integr1}\end{equation}
where the $D$ matrix defined in (\ref{DMATRIX}) is written as $D=\left[\begin{array}{cc}
A_{11} & - A_{12}\\
- A_{12} & A_{22}\end{array}\right]$, $A_{11}=I$, $A_{12}=\left[\begin{array}{cc}
1 & 0\\
0 &  C'\end{array}\right]$ , $A_{22}=\left[\begin{array}{cc}
1+\frac{1}{\sigma_{*}^{2}} & -\frac{{\bf a}^{T}}{\sigma_{*}^{2}}\\
-\frac{{\bf a}}{\sigma_{*}^{2}} & I+\Sigma^{-1}+\frac{{\bf a}{\bf a}^{T}}{\sigma_{*}^{2}}\end{array}\right]$, and $k_{3}=(2\pi)^{-\frac{N+1}{2}}|A_{22}|^{-\frac{1}{2}}$$|D|^{\frac{1}{2}}$.
Some manipulations lead to:

\begin{eqnarray*}
I_{1}&=&k_{3}\int\limits _{{\bf v}\ge0}\int\limits _{-\infty\le {\bf v}'\le\infty}\frac{e^{-\frac{1}{2}\left({\bf v}'-A_{22}^{-1}A_{12}{\bf v}\right)^{T}A_{22}\left({\bf v}'-A_{22}^{-1}A_{12}{\bf v}\right)}}{(2\pi)^{\frac{N+1}{2}}|A_{22}|^{-\frac{1}{2}}}d{\bf v}'\\
&.& e^{-\frac{1}{2}\left({\bf v}^{T}A_{11}{\bf v}-{\bf v}^{T}A_{12}A_{22}^{-1}A_{12}{\bf v}\right)}d{\bf v}\label{eq_Integr1_mod}\end{eqnarray*}
Now the inside integral w.r.t. ${\bf v}'$ equals 1. We get

\begin{equation}
I_{1}=k_{3}(2\pi)^{\frac{N+1}{2}}|A|^{-\frac{1}{2}}\int\limits _{{\bf v}\ge0}^{}{\cal N}\left({\bf v};0,A^{-1}\right)d{\bf v},\label{eq_Integr1_Final2}\end{equation}
where \begin{equation}
A=I-A_{12}A_{22}^{-1}A_{12}\label{eq_A}\end{equation}
A similar formula applies for $I_{2}$ but with integration limits
given by $-\infty<v_{1}<\infty$ and $\left[\begin{array}{c}
v_{2}\\
\vdots\\
v_{N+1}\end{array}\right]\ge0$

Some further simplification (see the Appendix for a detailed proof) leads to
\begin{eqnarray}\label{A-INV-FIN}
A^{-1}\equiv R &=& \pmatrix{ {1+\Sigma_{{\bf x}_* {\bf x}_*}} & 
{\Sigma_{X {\bf x}_*}^T C'} \cr
{C' \Sigma_{X {\bf x}_*}} & 
C'(I+\Sigma )C' \cr }  \label{A-INV-FIN1}\\
&=& I+ A_{12}\Sigma' A_{12} \label{A-INV-FIN2}
\end{eqnarray}
where $\Sigma'$ is the composite covariance function
(of training patterns plus testing pattern, see Eq. (\ref{F-FSTAR})):
\begin{equation}\label{COMP-COV}
\Sigma' =
\pmatrix{ {\Sigma_{{\bf x}_* {\bf x}_*}} & 
{\Sigma_{X {\bf x}_*}^T } \cr
{\Sigma_{X {\bf x}_*}} & 
\Sigma  \cr } 
\end{equation}
The final classification posterior probability is thus given by
\begin{equation}\label{J-FIN}
{J_{*}=p(y_*=1|X,{\bf y},{\bf x}_{*})=\frac{\int_{orth}{\cal N}\left({\bf v};0,I+ A_{12}\Sigma' A_{12}\right)d{\bf v}}{\int_{orth+}{\cal N}\left({\bf v};0,I+ A_{12}\Sigma' A_{12}\right)d{\bf v}}=\frac{I_{1}}{I_{2}}}\end{equation}
where $orth$ means the integration over ${\bf v} \ge 0$, and $orth^+$ is the integration over the region
given by $-\infty<v_{1}<\infty$ and $\left[\begin{array}{c}
v_{2}\\
\vdots\\
v_{N+1}\end{array}\right]\ge0$
As we have demonstrated in Eq. (\ref{PYX}), the term $p({\bf y}|X)$, which represents
the marginal likelihood as described in Eq. (\ref{MARG-LIK}) and beyond, is basically $I_2$, or the denominator
of the expression for $J_*$. Thus,
\begin{eqnarray}
L=p({\bf y}|X)&=&\int_{orth+}{\cal N}\left({\bf v};0,I+ A_{12}\Sigma' A_{12}\right)d{\bf v} \label{MARG-LIK-FIN}\\
&=&\int_{orth}{\cal N}\left({\bf v}';0,C'(I+ \Sigma')C'\right)d{\bf v}' \label{MARG-LIK-FIN2}
\end{eqnarray}
The last identity is obtained by noting that the limit for $v_1$ in Eq. (\ref{MARG-LIK-FIN}) 
stretches from $-\infty$ to $\infty$ and therefore $v_1$ can be integrated out.
The vector ${\bf v}'\equiv (v_2,\dots, v_{N+1})^T$ corresponds to
the variables of the training set, while $v_1$ corresponds to the test pattern.

\subsection{Some Insights}
The denominator, as mentioned, represents the marginal likelihood $p({\bf y}|X)$.
Essentially, the marginal likelihood,
as also  observed from the integral,
measures how well the class memberships $y_i$
fit with the covariance structure.
If the patterns of 1's and -1's multiplied with the components of $\Sigma'$ (through $A_{12}$)
emphasize the large covariance elements (through agreeing class memberships, i.e. $y_i y_j=1$)
and deemphasizing the small covariance elements (through disagreeing class memberships, i.e. the multiplied
by the factor $y_i y_j=-1$), then we have a relatively large marginal likelihood,
and therefore a fairly consistent model. The reason why the positive orthant integral
will have higher value for large positive covariances (rather than negative) is that the 
largest principal components will then be oriented closer to the orthant.

The classification posterior probability $J_*$
is the ratio of the integral over an orthant, and
the integral is over $orth+$ which basically extends
over two orthants. This makes the expression appropriately
smaller than 1. Observing the numerator, we find
that the predominant signs (in $y_i$ and therefore also $A_{12}$)
that multiply large covariance elements (in $\Sigma_{X{\bf x}_*}$)
will determine if the numerator has a large value (therefore
the pattern should be classified as Class 1), or a small value
(therefore
the pattern should be classified as Class 2). Note that the identity
matrix component of the covariance matrix in Eq. (\ref{A-INV-FIN2}) is some
kind of regularizing factor that acts to ``fatten"
density.

Let us now consider the effects of the hyperparameters.
Consider the RBF kernel (Eq. (\ref{RBF})). The analysis here complements and in many ways confirms 
the insightful analysis of Nickisch and Rasmussen \cite{NICKISCH08}.
When either the latent function scale  $\beta\longrightarrow 0$
 or the length scale $\alpha\longrightarrow 0$
 then the covariance matrix in the multivariate Gaussian
 tends to the identity matrix.
 In case of identity matrix the orthant
 integral equals $2^{-d}$ where $d$ is
 the dimension. So, we end up with $L=2^{-N}$
 and $J_*={1\over{2}}$.
 If $\beta\longrightarrow \infty$ then
 the identity matrix part of the covariance matrix becomes negligible compared to the other part,
 see Eq. (\ref{A-INV-FIN2}), and we get
a formula similar to Eq. (\ref{J-FIN}), but without the identity matrix added in the covariance
expression.
If $\alpha\longrightarrow \infty$, then we get the following formula.
To avoid distraction to side issues, the proof is not given here. 
\begin{equation}
J_*={{\int_{-\infty}^{\infty}e^{-{{u^2}\over{2}}}\sigma\bigl(\sqrt{\beta}u\bigr)^{N_1+1}\Bigl[1-\sigma\bigl(\sqrt{\beta}u\bigr)\Bigr]^{N_2}du}\over{\int_{-\infty}^{\infty}e^{-{{u^2}\over{2}}}\sigma\bigl(\sqrt{\beta}u\bigr)^{N_1}\Bigl[1-\sigma\bigl(\sqrt{\beta}u\bigr)\Bigr]^{N_2}}du}
\end{equation}
where $\sigma$ is the cumulative Gaussian integral (i.e. the integral of
 the one-dimensional Gaussian density), and $N_1$ ($N_2$) is the number of Class 1 (Class 2) training patterns.
 Essentially, this gives some kind of a ``soft" counting procedure,
 without regards to the distances involved. One can see
 that all test patterns will be classified as only one specific class (the one
that wins the counting game).

\section{The Multivariate Gaussian Integral}
As can be seen the final equations (\ref{J-FIN}), (\ref{MARG-LIK-FIN}), and (\ref{MARG-LIK-FIN2})  for the posterior probability
and the marginal likelihood are given in terms of two multivariate Gaussian integrals.
The difficulty is that these are very high dimensional integrals (the dimension equals
the size of the training set),
making that a formidable problem. Consider that we would like to
apply the standard approach of generating many points according to
the multivariate density, and then computing the fraction of points that fall
in the considered orthant (area of integration).  Such high order 
integrals are typically a very small number, for example $10^{-50}$.
So  even if we generate trillions of points, essentially no point will happen 
fall in the area of integration.

There is a large literature on the multivariate Gaussian integral.
Interest in the problem started around the forties 
 of last century, and research is continuing since then,
 see the reviews of Gupta \cite{GUPTA63a} and \cite{GUPTA63b}, Johnson \cite{JOHNSON94}, and Genz \cite{GENZ09}. Essentially the state of
 the art is that a closed form formula exists only for a dimension up to three
 for the centered case (i.e. the integrals for the zero mean 
case where the limits are from 0, as is our case, see Eriksson \cite{ERIKSSON90}), and no closed form formulas exist
 for the non-centered case. There are some special constructions of the covariance 
matrices for which simplified formulas exist (for any arbitrary dimension).
There has also been some series expansions for the centered case in terms of
the elements of the covariance matrix (Kendall \cite{KENDALL41}, and Moran \cite{MORAN48}), or in terms of the elements
of the inverse of the covariance matrix (Ribando \cite{RIBANDO06}). These formulas,
while very elegant and insightful, are intractable for dimensions larger than ten,
because of the huge number of combinations of powers of the $N^2$ variables of the covariance matrix.
A parallel track in the attempt to tackle  the multivariate Gaussian integral is
by applying some efficient numerical integration techniques (see for example Schervish \cite{SCHERVISH84}). 
Due to the exponential nature of these methods, they  are applicable 
for a dimension up to around 20.
Another track considers improvised Monte Carlo methods
(De\'ak \cite{DEAK86}, Genz \cite{GENZ93}, Breslaw \cite{BRESLAW94} and Hajivassiliou et al \cite{HAJIVASSILIOU91}).
Again, these methods have not been demonstrated on high dimensional problems.

\section{New Monte Carlo Method}
\subsection{The Proposed Method}\label{ALG1}
The proposed new Monte Carlo method tackles
the high-dimensional multivariate Gaussian integral,
and thereby simultaneously evaluates the posterior probability 
and the marginal likelihood. 
It combines aspects 
of rejection sampling and bootstrap sampling.
The general idea is to first generate samples 
for the first variable $v_1$. Subsequently,
we reject the points that fall outside the integral limits
(for $v_1$). Then we replenish in place of the discarded
points by sampling with replacement from the existing points (i.e. the points that have been accepted).
Next, we move on to the second variable, $v_2$, and generate
points using the conditional distribution $p(v_2|v_1)$.
Again, we reject the points of $v_2$ that fall outside
the integration limit, and replenish by sampling with replacement.
We continue in this manner until we reach the final variable
$v_N$. The integral value is then estimated as the product
of the acceptance ratios of the $N$ variables.
Unlike MCMC-type methods, we do not need 
to perform additional cycles. We cycle only 
once through the $N$ variables, each time generating
a number $M$ of points.

Here are the detailed steps of the algorithm:

\begin{enumerate}
\item For $i=1$ to $N$ perform the following:
\item 
If $i=1$ then generate $M$ points $v_1(m)$ from $p(v_1)$.
Otherwise, generate $M$ points $v_i(m)$ according to the conditional
density function:

\begin{equation}\label{GEN-DIST}
 p\equiv p(v_i|{\bf v}_{1:i-1}(m))
 \end{equation}
where ${\bf v}_{1:i-1}(m)$ is the $m^{th}$ string of points (of variables $v_1$ to $v_{i-1}$) already generated
in the previous steps.
\item
Reject the points $v_i(m)$ that are outside the area
of integration, i.e. reject the points 
$v_i(m)\le 0$. Assume that there are $M_1(i)$ accepted points
and $M_2(i)\equiv M-M_1(i)$ rejected point.
\item
Replenish in place of the rejected points, by 
sampling a number $M_2(i)$ points by replacement from among
the accepted points.
\item
Once reaching the last dimension $i=N$, we stop, computing
the multivariate integral as 
\begin{equation}\label{RATIO-MS}
I=\prod_{i=1}^N \Biggl( {{M_1(i)}\over{M}} \Biggr)
\end{equation}
\end{enumerate}
\subsection{The Rationale of the Algorithm} 
The proof that the proposed algorithm leads to
an estimate for the multivariate orthant probability
is essentially by construction, and this is described here.
The multivariate integral can be written as:
\begin{eqnarray}
I&=&p(v_1\ge 0, v_2\ge 0,\dots, v_N\ge 0) \\\label{RATIONALE1a}
&=& p(v_N\ge 0 | v_{1}\ge 0,\dots ,v_{N-1}\ge 0)\dots p(v_2\ge 0 |v_1\ge 0)p(v_1\ge 0) \label{RATIONALE1b}
\end{eqnarray}
Since we generate points according to the distribution in (\ref{GEN-DIST}),
which conditions only on the surviving points that were not being
rejected in previous rounds, the generated points will obey the
distribution $p(v_i|v_1\ge 0 ,\dots , v_{i-1}\ge 0)$.
As such, the ratio $M_1(i)/M$ is an estimate of the probability 
$p(v_i\ge 0 |v_1\ge 0 ,\dots , v_{i-1}\ge 0)$. Using Eq.(\ref{RATIONALE1b})
we obtain the product formula (\ref{RATIO-MS}) for the overall multivariate integral.
Please note that the bootstrap sampling step does not alter the distribution
of the generated points. 
\vskip .1in
\noindent
\subsection{Example:}
Consider for example that we would like to compute
\begin{equation}
I=\int_{0.7}^\infty \int_{1.4}^\infty \int_{1.2}^\infty {\cal N}(\mu,\Sigma) dv_1 dv_2 dv_3
\end{equation}

\begin{enumerate}
\item
We generate $M$ points from $p(v_1)$ (let $M=10$).
Let these points be 2.1, 0.2, 0.6, 1.8, 2.2, 0.8, 1.3, -0.3, 1.4, 1.6.
We reject the points 0.2, 0.6, 0.8, -0.3, as they are 
less than 1.2 and hence are outside the area of integration.
We keep the six accepted points 2.1, 1.8, 2.2, 1.3,  1.4, 1.6,
and remove the rejected points. In place of the rejected
points, we sample a similar number (i.e. four) from among
the accepted points, with replacement.
Assume we have done that and we have obtained: 2.1, 1.3, 1.8, 2.1.
So, overall we have the following points ($v_1(m)$'s):
 2.1, 1.8, 2.2, 1.3,  1.4, 1.6, 2.1, 1.3, 1.8, 2.1.
 \item
Generate $M\equiv 10$ points from $p(v_2|v_1(m))$.
Specifically, we generate one point $v_2(1)$ using
the density $p(v_2|v_1=2.1)$, then one point $v_2(2)$
using the density $p(v_2|v_1=1.8)$, and so on.
Let these generated points be
$v_2(m)=$ 2.5, 1.5, 2.0, 1.7, 1.1, 1.5, 1.7, 1.3, 1.9, 2.0.
We reject the points 1.1 and 1.3 as they are below the integration
limit of 1.4. Then we sample two more points by replacement
in place of these rejected points. We continue in this manner for the remaining dimension.
Assume that we rejected three points in this case.
\item The integral estimate is the product of the acceptance ratios, i.e.
it equals $\Bigl(\frac{6}{10}\Bigr)\Bigl(\frac{8}{10}\Bigr)\Bigl(\frac{7}{10}\Bigr)$.
\end{enumerate}

\subsection{On the Convergence of the Proposed Algorithm}


Because the integral is typically a very small number,
we will consider here the logarithm of the integral as the target value we
would like to estimate, i.e., from Eq.(\ref{RATIONALE1a}):
\begin{equation}\label{LOG-I}
{\rm log}(I)={\rm log}\Bigl[p(v_N\ge 0 | v_{1}\ge 0,\dots ,v_{N-1}\ge 0)\Bigr]+\dots + {\rm log}\Bigl[p(v_2\ge 0 |v_1\ge 0)\Bigr]
+{\rm log}\Bigl[p(v_1\ge 0)\Bigr]
\end{equation}
Assume for the time being that the values generated are independent.
In every step of the algorithm we generate $M$ Bernoulli trials, where each
has a probability of $P_i\equiv p(v_i\ge 0 |v_1\ge 0,\dots, v_{i-1}\ge 0)$ in 
landing in the integral's sought interval and being accepted for the subsequent steps.
Let us analyze the bias and the variance.
\begin{equation}\label{BIAS}
E\Biggl[{\rm log}\Biggl({{M_i}\over{M}}\Biggl)\Biggr] \approx {\rm log}\bigl(P_i\bigr)+{1\over{P_i}}E\Biggl[
{{M_i}\over{M}}-P_i\Biggr]- {{E \Biggl[{{M_i}\over{M}}-P_i\Biggr]^2}\over{2P_i^2}}
\end{equation}
where the expression in the RHS originates from a Taylor series expansion 
of log$(M_i/M)$ around the value log$(P_i)$, and keeping up to quadratic terms.
The expectation in the second term in the RHS equals zero (because it is a binomial process, so 
the expectation of $M_i$ equals $MP_i$).
The last term in the RHS can also be evaluated, and we obtain the bias as

\begin{eqnarray}\label{}
{\rm Bias} & \equiv & E\Biggl[{\rm log}\Biggl({{M_i}\over{M}}\Biggl)\Biggr]-{\rm log}\bigl(P_i\bigr) =-{{(1-P_i)}\over{2MP_i}}+O(M^{-2})\\
            &=& O(M^{-1})
\end{eqnarray}
which means that we can have the bias as close as possible to zero, as
the number of generated points becomes very large.

Concerning the variance, we get
\begin{equation}\label{VAR1}
{\rm Var}
\Biggl[{\rm log}\Biggl({{M_i}\over{M}}\Biggl)\Biggr]=
{{1-P_i}\over{MP_i}}+O(M^{-2}) = O(M^{-1})
\end{equation}
For the overall integral, we get
\begin{equation}\label{INT30}
{\rm log}(I)={\sum_{i=1}^N} {\rm log} (P_i)
\end{equation}
with the bias and variance becoming
\begin{equation}\label{BIAS2}
{\rm Bias}(\hat{I}) = -{1\over{M}} \Biggl[{\sum_{i=1}^N}{{(1-P_i)}\over{2P_i}}\Biggr] +O(M^{-2})
\end{equation}
\begin{equation}\label{VAR2}
{\rm Var}(\hat{I})={1\over{M}} \Biggl[{\sum_{i=1}^N}
{{(1-P_i)}\over{P_i}} \Biggr] +O(M^{-2})
\end{equation}
Thus, the mean square error (MSE) goes to zero as $M\longrightarrow\infty$. 
Note that $P_i$ is the outcome of a one-dimensional integral,
so it is expected to be in the middle range of $(0,1)$.

As a benchmark comparison, consider  the basic Monte Carlo integration
algorithm, where we generate a number of points according to the multivariate
Gaussian distribution and evaluate the fraction of points
falling in the area of integration. In that case
the mean square error is $(1-I)/(M I)+O(M^{-2})$,
where $I$ is the value of the multivariate integral.
One can see that the MSE is very large because typically
$I$ is infinitessimally small.

When we derived the above formula, we assumed that
the samples are independent, as an approximation. Strictly speaking they are not,
because of the following reason. Consider two
points $v_j(1)$ and $v_j(2)$ generated according to 
$p(v_j | v_{1}\ge 0,\dots ,v_{j-1}\ge 0)$.
Tracking backwards from their values at dimension $j$
and going upstream through the conditioned variables, we could find 
one variable, say $v_{j-k}$, that is
a common conditioned variable to the two
generated points $v_j(1)$ and $v_j(2)$ (i.e. a common ancestor).
This is because of the bootstrap sampling procedure.
However, we argue that the dependence will
be fairly small. This is because equally-valued samples 
will get completely dispersed when we generate samples
for the next variable, and so the dependence will decay fast. So, the net effect of this dependence
is to have a somewhat higher MSE, but it would
still be the same order, i.e. $O(M^{-1})$, and with
higher coefficient. (It is akin to estimating the mean 
of a variable using generated points having a banded covariance matrix,
the MSE will still be  $O(M^{-1})$.)

 \subsection{On Generating from the Distribution $ p(v_i|{\bf v}_{1:i-1}(m))$}
 In Step 2 in the algorithm described in Subsection \ref{ALG1}, we need to generate from the 
 conditional Gaussian distribution. This can be accomplished 
 using the well known identity (assume mean(${\bf v}$)=0):
 
\begin{equation}\label{V-COND}
p(v_i|{\bf v}_{1:i-1})= {\cal N}\Bigl( v_i,{\bf b}_i^T {\bf v}_{1:i-1},
\sigma_i^2 
\Bigr)
\end{equation}
where
\begin{equation}\label{V-MEAN}
{\bf b}_i=R^{-1}_{1:i-1,1:i-1}R_{1:i-1,i}
\end{equation}

\begin{equation}\label{V-VAR}
\sigma_i^2=R_{i,i}-R_{i,1:i-1}R^{-1}_{1:i-1,1:i-1}R_{1:i-1,i}
\end{equation}
and where $R\equiv I+{A_{12}}\Sigma'{A_{12}}$ is the covariance matrix
pertaining to the multivariate Gaussian (see Eqs. (\ref{A-INV-FIN1}) and (\ref{J-FIN})),
and the notation $A_{i:j,k:l}$ means the submatrix constructed from $A$ by taking
rows $i$ to $j$ and columns $j$ to $k$.

We have to compute these variables in Eqs. (\ref{V-MEAN}) and (\ref{V-VAR}), including inverting a matrix
every step, i.e. $N$ times. We present here a computationally
more efficient algorithm based on a recursive computation 
of the quantities in Eqs. (\ref{V-MEAN}) and (\ref{V-VAR}).
Assume that we have performed the computations
at Step $i$, i.e. that ${\bf b}_{i}$, $\sigma_{i}^2$ and $Q_{i}\equiv R^{-1}_{1:i-1,1:i-1}$
are available.
Proceeding to the next step  $i+1$, we first tackle $Q_{i+1}$. Using the 
partitioned matrix inversion (Horn and Johnson \cite{HORN85}), we get

\begin{eqnarray}\label{MAT-PART}
Q_{i+1} &=&\pmatrix{R_{1:i-1,1:i-1} &  R_{1:i-1,i} \cr
R_{1:i-1,i}^T & R_{ii}
             }^{-1}  \\
             &=& \pmatrix{R^{-1}_{1:i-1,1:i-1}+ {1\over{k}}R^{-1}_{1:i-1,1:i-1} R_{1:i-1,i}R^T_{1:i-1,i}R^{-1}_{1:i-1,1:i-1}
             & \ & -{1\over{k}}R^{-1}_{1:i-1,1:i-1} R_{1:i-1,i} \cr
            - {1\over{k}} R^T_{1:i-1,i}R^{-1}_{1:i-1,1:i-1} & \ & {1\over{k}}
             } \nonumber
\end{eqnarray}
where $k=R_{ii}-R_{1:i-1,i}^TR^{-1}_{1:i-1,1:i-1} R_{1:i-1,i}$.
Notice that $k$ equals $\sigma^2_i$, which is available from the previous step.
This way of updating the inverse of the covariance matrix has been commonly used
in the signal processing community, and it was even introduced in Gaussian process regression by
Csat\'o and Opper \cite{CSATO02}, Van Vaerenbergh et al \cite{VAEREN12}, and P\'erez-Cruz et al \cite{PEREZ13}.
Substituting from Eq. (\ref{V-MEAN}), we get
\begin{equation}\label{MAT-PART2}
Q_{i+1} = \pmatrix{Q_i+ {{\bf b}_i {\bf b}_i^T\over{\sigma^2_i}}  
             & \ & - {{\bf b}_i\over{\sigma^2_i}} \cr
           - {{\bf b}_i^T\over{\sigma^2_i}} & \ & {1\over{\sigma^2_i}}
             }
\end{equation}
We also get 
\begin{equation}\label{B-NEXT}
{\bf b}_{i+1}=Q_{i+1}R_{1:i,i+1}
\end{equation}

\begin{equation}\label{SIG-NEXT}
\sigma_{i+1}^2=R_{i+1,i+1}-R_{1:i,i+1}^T {\bf b}_{i+1}
\end{equation}
In summary, using these recursive formulas we can compute
the moments for the conditional distribution
using $O(N^2)$ instead of $O(N^3)$ operations, thus
providing some computational savings.

\subsection{Summary of the Algorithm}
The algorithm turns out to be very simple,
and can be coded easily.
It is important to start with the training set,
then proceed with the test set. So, basically
we will rename the variables, such that
${\bf v}_{1:N}$ represents the training set,
and ${\bf v}_{N+1:N+NTEST}$ represents the test set.
Also, for convenience, the covariance matrix of (\ref{A-INV-FIN1}) will be rearranged
and will be made to include all test patterns,
to become
\begin{equation}
 R = \pmatrix{ C'(I+\Sigma )C' &  
{\Sigma_{X X_*}^T C'} \cr
{C' \Sigma_{X X_*}} & 
{I+\Sigma_{X_* X_*}} & \cr }  \label{R-NEW}\\
\end{equation}
where $X_*$ is the matrix of test patterns. As can be seen in the algorithm,
the training set computations have to be done once,
and need not be repeated for every test pattern, making the algorithm of incremental
nature. The algorithm is described follows:
\vskip .1in\noindent
{\bf Algorithm GPC-MC}
\begin{enumerate}
\item $i=1$:
Set $Q_2=\frac{1}{R_{1,1}}, \sigma_1^2=R_{1,1}$. Generate $M$ points $v_1(m)$ according
to ${\cal N}(v_1,0,\sigma_1^2)$. Compute 
\begin{equation}\label{FRAC-V1}
\hat{P_1}=\frac{\#\Bigl(v_1(m)\ {\rm s.\ t.\ }v_1(m)\ge 0\Bigr)}{M}
\end{equation}
where the latter expression means the fraction of points
that are $\ge 0$. Remove the points $v_1(m)<0$. Sample by replacement
from among the remaining points to keep the total number
of points equal $M$. Rename the variables, so that $v_1(m)$
are the new kept points.
\item For $i=2$ to $N$ do the following:
\begin{enumerate}
\item Compute the matrices:
\begin{equation}\label{B-NEXT-A1}
{\bf b}_{i}=Q_{i}R_{1:i-1,i}
\end{equation}

\begin{equation}\label{SIG-NEXT-A1}
\sigma_{i}^2=R_{i,i}-R_{1:i-1,i}^T {\bf b}_{i}
\end{equation}
\begin{equation}\label{MAT-PART-A1}
Q_{i+1} = \pmatrix{Q_i+ {{\bf b}_i {\bf b}_i^T\over{\sigma^2_i}}  
             & \ & - {{\bf b}_i\over{\sigma^2_i}} \cr
           - {{\bf b}_i^T\over{\sigma^2_i}} & \ & {1\over{\sigma^2_i}}
             }
\end{equation}

\item Generate $M$ points $v_i(m)$ according
to ${\cal N}(v_i,{\bf b}_i^T{\bf v}_{1:i-1}(m),\sigma_i^2)$, $m=1,\dots, M$. Compute 
\begin{equation}\label{FRAC-VI}
\hat{P_i}=\frac{\#\Bigl(v_i(m) \ {\rm s.\ t.\ } v_i(m)\ge 0\Bigr)}{M}
\end{equation}
\item Remove the points $v_i(m)<0$. In their place, sample by replacement
from among the remaining points to keep the total number
of points equal $M$. Rename the variables, so that $v_i(m)$
are the new kept points. 
\end{enumerate}
\item The log marginal likelihood function
is given by the following sum over the training set probabilities:
\begin{equation}
LogL={\sum_{i=1}^N} {\rm log}\bigl(\hat{P_i}\bigr)
\end{equation}
\item For $i=N+1$ to $N+NTEST$ (the test patterns) do the following:
\begin{enumerate}
\item Compute the matrices:
\begin{equation}\label{B-NEXT-A2}
{\bf b}_{i}=Q_{N+1}R_{1:N,i}
\end{equation}
where $Q_{N+1}$ represents the  
covariance matrix inverse, obtained
at the last training pattern. It will not be
be updated further during the test.
\begin{equation}\label{SIG-NEXT-A2}
\sigma_{i}^2=R_{i,i}-R_{1:N,i}^T {\bf b}_{i}
\end{equation}
\item Generate $M$ points $v_i(m)$ according
to ${\cal N}(v_i,{\bf b}_i^T{\bf v}_{1:N}(m),\sigma_i^2)$, $m=1,\dots, M$. Compute 
\begin{equation}\label{FRAC-VI-2}
\hat{P_i}=\frac{\#\Bigl(v_i(m) \ {\rm s.\ t.\ } v_i(m)\ge 0\Bigr)}{M}
\end{equation}
Note that $\hat{P_i}$ is the sought {\it test pattern 
posterior probability}. Note also that we do not need to perform the bootstrap sampling step
here for the test.
\end{enumerate}
\end{enumerate}
\vskip .2in\noindent
Note that the proposed algorithm, after it is applied to the training set,
has its samples obey the posterior distribution. So these samples
can be saved for any future evaluation of a test pattern.

 \subsection{Another Variant}\label{SOFT-COUNT}
 A possibly more efficient modification is to have some kind of soft count, instead of the
 hard count used in  Eq.(\ref{FRAC-VI}) (or Eq. (\ref{RATIO-MS}) ). We know that each point, while moving
 from Step $i-1$ to Step $i$, is generated from a Gaussian density. The probability
 of landing in the positive side can be computed by simply applying the cumulative Gaussian integral for each point
 (i. e. $\sigma (u) \equiv \int_{-\infty}^{u}\frac{e^{-\frac{x^{2}}{2}}}{\sqrt{2\pi}}dx$). In that case, 
 instead of Eq. (\ref{FRAC-VI}) we apply the following:
\begin{equation}\label{RATIO-MS-2}
\hat{P_i}=\frac{
\sum_{m=1}^M {\sigma\Biggl( \frac{{\bf b}_i^T {\bf v}_{1:i-1}(m)}{\sigma_i}\Biggr)}}{M}
\end{equation}
This applies similarly to Eq. (\ref{FRAC-VI-2}).
Of course, this does not relieve us from having to generate the points. We have to do that while moving forward
till reaching the last dimension. In essence, all other steps are similar to the original
version of the algorithm. Only the count is different.

\section{Simulation Experiments}
\subsection{Experiment 1: Testing the New Monte Carlo Integration Method}
In this experiment we test the convergence properties of the new Monte Carlo multivariate
Gaussian integration approach. The goal here is to test the efficacy of this new
method irrespective of its use in Gaussian process classifiers.
The application of this method to the Gaussian process classifier 
hinges mainly on its success as a stand-alone
integration approach. Once we establish this fact, we will have assurances 
that it would work well in the Gaussian process classifier setting.

To be able to judge the new method's approximation error, we have to
use examples where the ``ground truth", i.e. the real integral value, is known. We identified
a special form where this can be obtained. The experiments will be performed
on this special form, described below.

The covariance matrix equals 1 on the diagonal 
and equals $d_i d_j$ off the diagonal (at the $(i,j)^{th}$ position),
where ${\bf d}=(d_1,\dots, d_N)^T$ is some vector with $|d_i|<1$.  
In such a situation, the orthant probability can be reduced to a simple
one-dimensional integration, as follows:
\begin{equation} \label{PROBVAL}
\int_{0}^\infty {\cal N}({\bf v},{\bf \mu},\Sigma)d{\bf v} = {1\over{\sqrt{2\pi}}} \int _{-\infty}^{\infty} e^{-{{u^2}\over{2}}}
\prod_{i=1}^N  {\sigma} \Biggl( {{d_i u}\over{\sqrt{1-d_i^2}}}\Biggr) du
\end{equation}
where $\sigma$ is the cumulative Gaussian function (i.e. the one-dimensional
integration of the Gaussian density function).
This formula was proposed by 
Das \cite{DAS56}, Dunnet and Sobel \cite{DUNNET55}, and Ihm \cite{IHM59}.
 It has been also generalized to different forms by
Marsaglia \cite{MARSAGLIA63} and Webster \cite{WEBSTER70}, and also used in combination 
of Monte Carlo sampling by 
Breslaw \cite{BRESLAW94}. Some special cases of this formula even yield some closed-form 
solutions.

In this experiment we considered 
different dimensions for our space. Specifically,
we considered the dimensions $N=50$, $N=200$, and $N=500$.
In addition, for each dimension we considered
50 different problems, where each problem has a different 
${\bf d}$ vector (whose components are generated from a uniform distribution in $[-1,1]$). 
To evaluate how the approximation error varies
with the number of Monte Carlo samples $M$,
we ran each of these problems for various values
of $M$. Because the value of the integral is usually
an infinitessimal value, a sensible approach is to
consider the logarithm of the estimated integral, and compare it to the logarithm of 
the true integral. For example a typical integral value
for a 500-dimensional problem could be $10^{-200}$. The logarithm
becomes about -461. 
As an error measure, we used the following {\it mean absolute percentage error},
defined as:
\begin{equation}\label{MAPE}
MAPE={100\over{NR}}{\sum_{i=1}^{NR}} {{|{\rm log}(I)-{\rm log}(\hat{I})|}\over{|{\rm log}(I)|}}
\end{equation}
where $I$ is the true integral value, $\hat{I}$ is the integral value estimated
by the algorithm, and $NR$ represents the number of runs (i.e. the number
of different ${\bf d}$ vectors tested, in our case $NR=50$).

Note that we have to be careful when evaluating the true integral
numerically using Eq. (\ref{PROBVAL}). If we multiply the terms
first and then integrate numerically, we end up with
very small numbers, leading to a large error. We overcame this 
difficulty, by successive normalization by the maximum value after each multiplication.
Then we evaluate the integral and multiply back the normalization terms that we divided by.

Table \ref{table1} shows the MAPE error measure (average over each of the 50 tested problems)
for each of the tested values of $N$ (dimension) and $M$ (number of Monte Carlo runs).
Note that these are {\it percent} errors, so they are multiplied by 100.
Displayed in the table is also the standard error (over the 50 tested problems).
One can observe that the developed algorithm evaluates the orthant probabilities with
good accuracy. As expected the accuracy tends to improve for larger $M$.
However, the relation between the accuracy and the dimension is less straightforward
to describe.  Even though by Eqs. (\ref{BIAS2}) and (\ref{VAR2}) one might expect
that for large $N$ there will be more terms and hence a higher error, 
in practice the $P_i$'s are more important influencing factors.
One can also see that the algorithm succeeded for even the case of  500 dimensional
problems, even though such high dimensions are quite formidable problems. Most of the algorithms
for orthant probability estimation test on problems with only tens of dimension.

Note that because of memory limitations, in case of a large number $M$ of Monte Carlo samples 
it may not be practical to propagate all samples together.
A more practical approach is to rerun the problem 
several times, each  with a smaller $M$.
For example assume that we would like
to use 2,000,000 samples. In that case
we  apply ten runs, each with $M=200,000$.
We then average the integral estimates obtained.
\begin{table}
\small
\begin{center}
\begin{tabular}{|c||c|c|c|c|c|c|}
\hline
No. MC samples	&	Problem 1 ($N=50$)		& Problem 2 ($N=200$)				&	Problem 3 ($N=500$)				\\
\hline\hline
3,000,000	&	0.012	(	0.0011	)	&	0.015	(	0.0065	)	&	0.043	(	0.0299	)	\\
1,000,000	&	0.016	(	0.0020	)	&	0.019	(	0.0068	)	&	0.045	(	0.0298	)	\\
300,000	&	0.041	(	0.0050	)	&	0.032	(	0.0072	)	&	0.050	(	0.0297	)	\\
100,000	&	0.072	(	0.0094	)	&	0.039	(	0.0070	)	&	0.059	(	0.0303	)	\\
30,000	&	0.141	(	0.0160	)	&	0.078	(	0.0065	)	&	0.080	(	0.0289	)	\\
10,000	&	0.245	(	0.0268	)	&	0.101	(	0.0093	)	&	0.107	(	0.0322	)	\\
\hline
\end{tabular} 
\end{center} 
\caption{The Mean Absolute Percentage Error (MAPE, in \%) of the Log Multivariate Gaussian Integral Estimate (and its Standard Error in Brackets) against 
the Dimension of the Problem and the Number of Monte Carlo Samples (the Numbers are in {\it Percent} so they are Multiplied by 100)} 
\label{table1}
\end{table}

\subsection{Experiment 2: Testing the New Monte Carlo Method on Gaussian Process Classification Problems}
The next group of experiments aims to verify that the proposed Monte Carlo method,
in a Gaussian process classification setting,
does converge to the true solution. The problem we face is that in general 
there is no way to know the true solution, and so it could be hard to verify 
this claim. However, we identified a special group of
problems where the ``ground truth" could be obtained.
This is if we take the distance kernel function to be of the dot product  form.
This means that the covariance matrix equals
\begin{equation}
\Sigma = X X^T 
\end{equation}
This is the so-called linear kernel. It is a legitimate kernel function as it represents a similarity between the patterns,
and is positive semidefinite.
For single-feature classification problems 
it can be shown 
after a few of lines of derivation that the class 1 posterior probability 
of a pattern, as given by Eq. (\ref{J-FIN}), becomes
\begin{equation}\label{ONE-D}
J_*={{
\int _{-\infty}^{\infty} e^{-{{u^2}\over{2}}}
\sigma \bigl( x u\bigr) \prod_{i=1}^{N} \sigma \bigl( y_i x_i u\bigr) du
}\over{
\int _{-\infty}^{\infty} e^{-{{u^2}\over{2}}}
\prod_{i=1}^{N} \sigma \bigl( y_i x_i u\bigr) du
}}
\end{equation}
where $x_i$ and $x$ are the feature values of respectively 
the $i^{th}$ training pattern and the test pattern, and $y_i$ denotes the class membership 
for pattern $i$. The marginal likelihood is simply the denominator in 
Eq. (\ref{ONE-D}). In this group of problems the covariance function becomes of the form discussed in the Experiment 1,
where we can make use of the one-dimensional integration method of Eq, (\ref{PROBVAL}) to evaluate the integrals.
The fact that we are dealing with a one-dimensional feature space does not necessarily make the problem any easier.
We are still dealing with the same formulas and with the same very high-dimensional integrals.
The dimension of the feature vector impacts only the covariance matrix, it will
just have different entries.

We generated a number of training and testing patterns from a one-dimensional (i.e. single-feature)
two-class Gaussian problem.
To ensure that the proposed model can handle different types of
problems, we considered a variety of training/testing set sizes, a variety
of means/variances for the class-conditional densities (in order to account
for a variety of different class overlaps).
Specifically, we considered the four problems shown below. Let $N$ and $NTEST$
be the sizes of respectively the training set and test set, and let $\mu_i$ and $\sigma_i$
be respectively the mean and standard deviation of the class conditional density for class $i$. 
\begin{itemize}
\item Problem 1: $N=100,\ NTEST=50,\ \mu_1=0,\ \mu_2=1,\ \sigma_1=0.2,\ \sigma_2=0.3$.
\item Problem 2: $N=200,\ NTEST=100,\ \mu_1=0,\ \mu_2=1,\ \sigma_1=2,\ \sigma_2=1$.
\item Problem 3: $N=400,\ NTEST=200,\ \mu_1=0,\ \mu_2=1.5,\ \sigma_1=0.5,\ \sigma_2=0.75$.
\item Problem 4: $N=800,\ NTEST=400,\ \mu_1=0,\ \mu_2=1,\ \sigma_1=1,\ \sigma_2=0.75$.
\end{itemize}
When constructing the covariance matrix, we made a point to 
shuffle the training patterns of both classes.
This is {\it important} for achieving better accuracies/speeds.
The reason will be mentioned at the end of this subsection.
To obtain statistically more reliable numbers,
from each of the above problems we applied the proposed algorithm a number $N_R\equiv 20$ different times.
Since the estimated class 1 posterior probabilities of the different test patterns  
are in a well-known range from 0 to 1, it is sufficient to use an absolute
error metric, so we used the mean absolute error (MAE), defined as follows:
\begin{equation}\label{MAE}
MAE={{1}\over{NTEST}}{\sum_{j=1}^{NTEST}} |J_{*j} -\hat{J}_{*j}|
\end{equation}
where $J_{*j}$ is an evaluation of the class 1 posterior probability for test pattern $j$
using an exact numerical integration procedure (the true value), obtained by the formula
of Eq. (\ref{ONE-D}),
and $\hat{J}_{*j}$ is the estimate using the proposed Monte Carlo procedure.
Table \ref{table2a} shows the obtained MAE values, averaged over the 20 runs, for a variety of numbers of Monte Carlo samples.

Concerning the log marginal likelihood, we evaluated it using the proposed
Monte Carlo algorithm,  and compared it
with the true value, obtained numerically by evaluating the denominator of Eq. (\ref{ONE-D}). Since the log marginal likelihood
can take any level, a normalized error measure is more appropriate.
So we used the mean absolute percent error (MAPE) measure. The formula is similar to
Eq. (\ref{MAPE}), but with the appropriate comparison
variables replaced.
Table \ref{table2b} shows the obtained MAPE (\%) values for a variety of numbers of Monte Carlo samples
for the log marginal likelihood estimation problem.

As seen from both tables, the algorithm is able to achieve a low error for both,
the probability evaluation and the marginal likelihood. One can also see that 
increasing the number of Monte Carlo samples $M$ leads to better accuracy.
As mentioned in the last experiment, the relation between accuracy
and dimension is less straightforward to describe. It is influenced more
by the specific covariance matrix and the resulting conditional
probabilities $P_i$. By observing 
Eqs. (\ref{BIAS2}) and (\ref{VAR2}), one finds that a small $P_i$
can lead to large error. It is therefore advantageous to 
have the $P_i$'s closer to the middle (in most cases it is around 0.5).
To achieve that, it is important to shuffle the data of both classes,
rather than list first the data for Class 1, followed by the data of Class 2.
The latter will cause more extreme $P_i$'s and therefore lead to less accuracy.
Other than random shuffle, one could interleave the data
of class 1 and class 2 in a repetitive way (e.g. class 1 pattern, then class 2, then class 1, then class 2, etc).
\begin{table}
\small
\begin{center}
\begin{tabular}{|c||c|c|c|c|c|c|}
\hline

No MC 	&	Prob 1 	&	Prob 2 	&	Prob 3 	&	Prob 4 	\\
Samples	&	($N=100$)	&	($N=200$)	&	 ($N=400$)	& ($N=800$)	\\
\hline\hline
3,000,000	&	0.00016	(	0.00004	)	&	0.00024	(	0.00005	)	&	0.00022	(	0.00005	)	&	0.00022	(	0.00006	)	\\
1,000,000	&	0.00031	(	0.00007	)	&	0.00043	(	0.00010	)	&	0.00038	(	0.00009	)	&	0.00045	(	0.00012	)	\\
300,000	&	0.00056	(	0.00013	)	&	0.00081	(	0.00018	)	&	0.00062	(	0.00014	)	&	0.00075	(	0.00019	)	\\
100,000	&	0.00095	(	0.00021	)	&	0.00139	(	0.00031	)	&	0.00112	(	0.00025	)	&	0.00128	(	0.00033	)	\\
30,000	&	0.00160	(	0.00036	)	&	0.00297	(	0.00066	)	&	0.00212	(	0.00047	)	&	0.00235	(	0.00061	)	\\
10,000	&	0.00308	(	0.00069	)	&	0.00463	(	0.00103	)	&	0.00391	(	0.00088	)	&	0.00443	(	0.00114	)	\\


\hline
\end{tabular} 
\end{center} 

\caption{The Mean Absolute Error (MAE) (averaged over the 20 runs) of the Gaussian Process Classification of Experiment 2 for the Four Different Problems against the Number of Monte Carlo
Samples (the Standard Error is in Brackets).} 
\label{table2a}
\end{table}

\begin{table}
\begin{center}
\begin{tabular}{|c||c|c|c|c|c|c|}
\hline

No MC	&	Prob 1 	&	Prob 2 	&	Prob 3 	&	Prob 4 	\\
Samples	&	($N=100$)	&	($N=200$)	&	 ($N=400$)	& ($N=800$)	\\
\hline\hline
	3,000,000	&	0.0081	(	0.0018	)	&	0.0088	(	0.0020	)	&	0.0063	(	0.0014	)	&	0.0033	(	0.0009	)	\\
	1,000,000	&	0.0187	(	0.0042	)	&	0.0097	(	0.0022	)	&	0.0095	(	0.0021	)	&	0.0059	(	0.0015	)	\\
	300,000	&	0.0364	(	0.0081	)	&	0.0255	(	0.0057	)	&	0.0130	(	0.0029	)	&	0.0077	(	0.0020	)	\\
	100,000	&	0.0671	(	0.0150	)	&	0.0340	(	0.0076	)	&	0.0238	(	0.0053	)	&	0.0130	(	0.0034	)	\\
	30,000	&	0.1170	(	0.0262	)	&	0.0629	(	0.0141	)	&	0.0450	(	0.0101	)	&	0.0249	(	0.0064	)	\\
	10,000	&	0.1522	(	0.0340	)	&	0.1334	(	0.0298	)	&	0.0900	(	0.0201	)	&	0.0622	(	0.0161	)	\\

\hline
\end{tabular} 
\end{center} 

\caption{The Mean Absolute Percent Error MAPE (in \%, i.e. the Numbers are Multiplied by 100) of the Log Marginal Likelihood of Experiment 2 for the Four Different Problems against the Number of Monte Carlo
Samples. All are Averages over the 20 Runs, and the Standard Error is in Brackets.} 
\label{table2b}
\end{table}


\subsection{Experiment 3: Comparison between the New Monte Carlo Method and the MCMC Approach}
In this and the next experiment we present a comparison of the proposed Monte Carlo algorithm with the Markov Chain Monte Carlo (MCMC) approach, its only peer.
The MCMC is the only available method that can accurately compute the exact classification probabilities. 
All other methods give only approximations.
There are several MCMC based models.
In this experiment we compare between the proposed algorithm and the Hybrid Monte Carlo (HMC) \cite{NEAL99}, and the Elliptical Slice Sampler (ESS) by Murray, Adams, and Mackay \cite{MURRAY10}. For both methods, we use the implementation
written by Rasmussen and Nickisch \cite{RASMUSSEN10}, which includes several enhancements of these two methods.
For the proposed algorithm we used the variant with the soft count, described in Subsection \ref{SOFT-COUNT}.

We considered  Problem 3 ($N=400$) of Experiment 2 (with a linear kernel). As mentioned, these
are the only type of problems where the ground truth is known.
For the purpose of comparison 
the two main aspects of speed and accuracy are important.
They are contradictory metrics, 
for example improving the accuracy (by having a larger Monte Carlo sample) will lead to more lengthy runs,
and vice versa too.
To be able to visualize simultaneously both of these aspects of the performance
we have plotted both the CPU time  against the logarithm
of the MAE in Figure 1 (for the case of probability estimation) and against the logarithm of the MAPE
in Figure 2 (for the case of marginal likelihood estimation).
In each of the two figures every point corresponds to the average CPU time/average MAE (or MAPE) over ten runs
for a particular Monte Carlo parameter. The Monte Carlo parameter for the proposed
algorithm, and for the HMC and ESS algorithms is the number of Monte Carlo samples. For the HMC and ESS algorithms 
we kept the other parameters at their recommended values, as given in the implementation
by Rasmussen and Nickisch \cite{RASMUSSEN10} (they are any way much less influential than the
number of samples). The parameters are fixed as follows:
the number of of skipped samples is 40, the number of burn-in samples is 10,
and the number of runs to remove finite temperature bias is 3.

To be able to judge the advantage of one algorithm versus another,
one should examine the difference in accuracy for the same
run time, or similarly the difference in run time for the same accuracy.
One can see from the graphs that the proposed algorithm 
generally beats the HMC algorithm. The margin of outperformance is considerable, especially
for the marginal likelihood case. The ESS ties with the the proposed algorithm for
pattern probability estimation for low to moderate accuracy targets, but ESS outperforms
for high accuracy computationally expensive runs. 
On the other hand, the proposed algorithm outperforms ESS considerably for the marginal likelihood case.
As pointed out before, the marginal likelihood is by far
the most important of the two aspects. The reason is that
it is evaluated numerous times in the process of tuning
the hyperparameters of the kernel function, while the pattern probability
estimation is performed only once.
For example, from the figure the proposed algorithm produces
a log(MAPE) of the marginal likelihood of about -4.52 with a CPU time 
of 525 sec. The ESS algorithm produces about the same log(MAPE) (or just a little better at -4.70) 
with CPU time of 5350 sec. With about a hundred application
of an optimization algorithm (such as Rasmussen and Nickisch's GPML toolbox's \texttt{minimize}  function \cite{RASMUSSEN10}),
the new algorithm takes
about 15 hours, while ESS takes about 148 hours.

\begin{figure}\label{MAE-FIG}
\begin{center}
\resizebox{3.5in}{!}{\includegraphics*{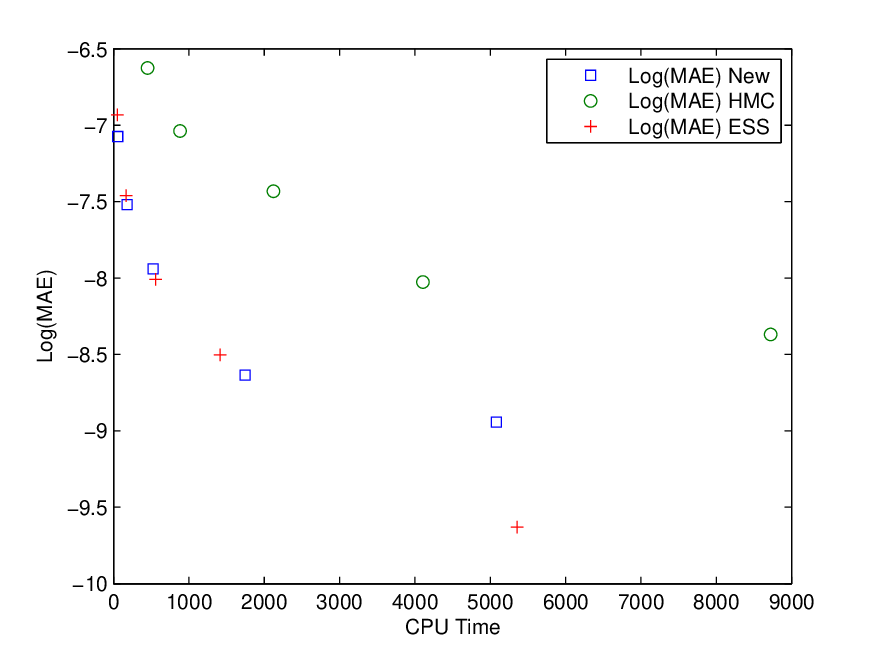}}
\caption{Log of the Mean Absolute Error (MAE) of the Probability Estimates 
of the New Algorithm, the HMC Algorithm, and the ESS Algorithm against the CPU Time (Seconds) of the Runs}
\end{center}
\end{figure}

\begin{figure}\label{MAPE-FIG}
\begin{center}
\resizebox{3.5in}{!}{\includegraphics*{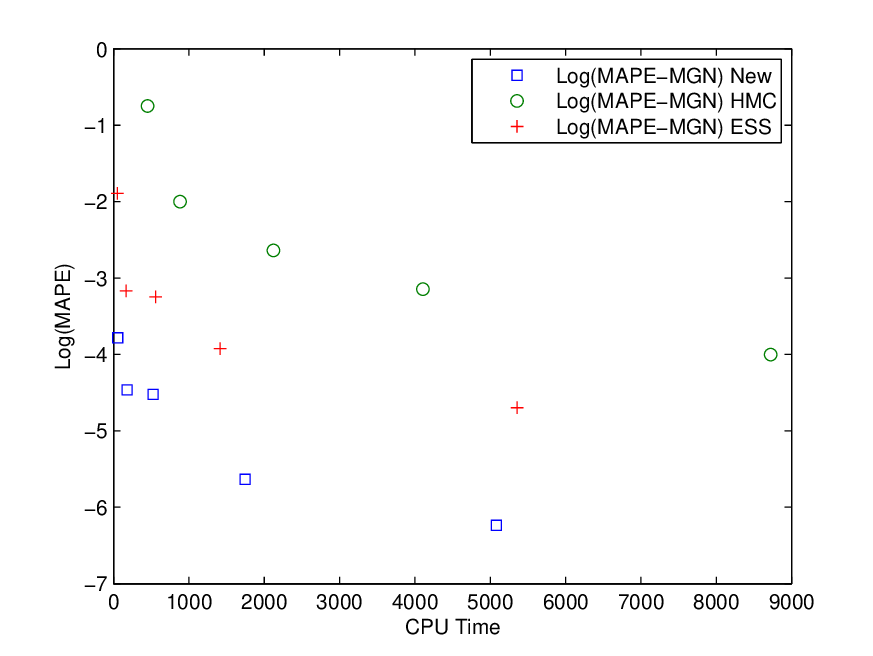}}
\caption{Log of the Mean Absolute Percentage Error (MAPE) of the Log Marginal Likelihood Estimates
of the New Algorithm, the HMC Algorithm, and the ESS Algorithm against the CPU Time (Seconds) of the Runs}
\end{center}
\end{figure}

\subsection{Experiment 4: Comparison with the Other MCMC Methods on Synthetic and Real World Problems}
The previous experiment, while providing an accurate
benchmark comparison, considers only the linear kernel,
which may not be very prevalent in real world applications.
In this experiment we consider the more common RBF kernel,
and also some synthetic and real world problems, in order to 
have a test as close as possible to realistic situations.
We also add to the comparison the model developed by Titsias, Lawrence and Rattray \cite{TITSIAS09},
in addition to the ESS and HMC algorithms considered last experiment.
Titsias et al's algorithm is another MCMC-based algorithm for the GPC problem.
It relies on using dynamically optimized control variables that provide a low
dimensional representation of the function. 
The code is publicly available at \cite{TITSIAS-CODE}.
It applies only to RBF and ARD kernels \cite{TITSIAS-CODE},
and that is why we did not include it in the comparison of the last experiment.

For the purpose of this comparison, there is however the problem of lack of ground truth,
because for RBF kernels we do not know the true value of the integrals that would yield
the Gaussian process class probabilities. 
Nevertheless, we run the competing models on
a number of problems, and check their convergence 
properties. If they converge to the same class probabilities,
then this is a strong indication that these algorithms
do indeed converge. Also, note that the considered RBF kernel
is a more efficient and more widely used kernel than the linear kernel.
So the experiments presented here are more relevant, as they fit more closely
to the real experimental situations.
 
We considered artificially generated data
using Gaussian class-conditional densities, and real data sets.
For the artificial problems, using the Bayes classifier's formula, one can compute the true posterior probability
$P(y_i=1|x)\equiv P^{true}_i$ which the Gaussian process classifier attempts to
model. However, we must emphasize that these true posterior probabilities
need not be the same as those obtained by GPC, as GPC is
based on a different formulation. 
If either proposed or competing algorithms do a good job converging
to the true value of the sought integral, but it turns out to be far
from the true posterior, then it is not the fault of the algorithm.
It should be attributed to the degree of validity of the Gaussian process 
formulation or to the finite sampled-ness of the training data.
Nevertheless,
a comparison with  the true posterior provides a useful sanity
check. We computed the mean absolute error 
between each competing method's estimated probabilities
and $P^{true}_i$ (let us denote them by  MAE-POST(NEW), MAE-POST(TITSIAS), MAE-POST(ESS), and MAE-POST(HMC)).
This measure applies only for synthetic problems, as for real problems
we do not know the true posteriors.
We also computed the following measure.
For each pattern, we obtain the median of all four algorithms' estimated
probability $P_i^{med}$.  This so-called ``consensus" value is compared
against each algorithm's estimated
probabilities. We get ${\rm Mean}_i\Bigl(|P_i-P_i^{med}|\Bigr)$ for each method
(denote these by MAE-MED(NEW), MAE-MED(TITSIAS), MAE-MED(ESS), and MAE-MED(HMC)).
This measure will expose the aberrant algorithm that fails to converge, and
is therefore  a useful sanity check. We have computed a similar measure
for the log marginal likelihood.

The problems considered are described as follows.
Let $d$ be the dimension of the feature vector, and let $e_d$ denote 
the $d$-dimensional vector of all ones. Also, let:
\begin{equation} 
 {\Sigma_0}=\pmatrix{1 & 0.25 \cr
                     0.25 & 1 \cr}
\end{equation}
and $\Sigma_{10}$ is a $10\times 10$ one-banded matrix 
with 0.5 on the diagonal and 0.2 on the upper and the lower bands.
 We considered the following synthetic problems, that provide a variety of different levels of 
 training set sizes, class overlaps, and space dimensions, and also the following real world problems.
 \begin{itemize}
 \item Problem 1: $NTRAIN=50,\ NTEST=50, \ d=2$, \ \ $p(x|C_1)={\cal N}(x,0,I)$, \ $p(x|C_2)={\cal N}(x,e_2,\Sigma_0)$.
 \item Problem 2: $NTRAIN=200,\ NTEST=200, \ d=2$, \ \ $p(x|C_1)={\cal N}(x,0,I)$, \ $p(x|C_2)={\cal N}(x,e_2,\Sigma_0)$.
 \item Problem 3: $NTRAIN=50,\ NTEST=50, \ d=2$, \ \ $p(x|C_1)={\cal N}(x,0,I)$, \ $p(x|C_2)={\cal N}(x,0.5 e_2,\Sigma_0)$.
 \item Problem 4: $NTRAIN=50,\ NTEST=50, \ d=2$, \ \ $p(x|C_1)={\cal N}(x,0,I)$, \ $p(x|C_2)={\cal N}(x,2e_2,\Sigma_0)$.
  \item Problem 5: $NTRAIN=50,\ NTEST=50, \ d=10$, \ \ $p(x|C_1)={\cal N}(x,0,I)$, \ $p(x|C_2)={\cal N}(x,e_{10},\Sigma_{10})$. 
 \item Problem 6: $NTRAIN=1000,\ NTEST=1000, \ d=2$, \ \ $p(x|C_1)={\cal N}(x,0,I)$, \ $p(x|C_2)={\cal N}(x,e_2,\Sigma_0)$.
\item Problem 7:  Crabs data, $NTRAIN=100,\ NTEST=100, \ d=6$, available at {\texttt http://www.stats.ox.ac.uk/pub/PRNN/}.
\item Problem 8:  Breast Cancer data, $NTRAIN=200,\ NTEST=249, \ d=9$, available at {\texttt http://mlearn.ics.uci.edu/databases/breast-cancer-wisconsin/}.
\item Problem 9:  USPS 3 vs 5 data, $NTRAIN=750,\ NTEST=790, \ d=256$, available at {\texttt http://www.gaussianprocess.org/gpml/data/}.
\end{itemize}
In the synthetic problems (1 to 6) we assume that the a priori probabilities are
equal.

We ran six different runs on each of these problems,
These six runs consider an RBF covariance
function $\Sigma$, with the following parameters:
\begin{enumerate}
\item $\alpha=5, \ \ \ \beta=1$
 \item $\alpha=5, \ \ \ \beta=5$
\item $\alpha=3, \ \ \ \beta=2$
\item $\alpha=0.5, \ \ \ \beta=0.5$
\item $\alpha=3, \ \ \ \beta=1$
\item $\alpha=0.5, \ \ \ \beta=3$ 
 \end{enumerate}
 For TITSIAS we considered 50,000 iterations, 
where we considered 5000 iterations for the burn-in,
 and a starting number of three control variables.
 For each problem we first ran the TITSIAS on the first $\alpha$ and $\beta$ parameter
 set. We selected the number of Monte Carlo samples of the proposed method, the ESS method, and the HMC method
 so that it runs in about the same time as the TITSIAS (measured by CPU time).
Then, we fixed this number for the other $\alpha$ and $\beta$ parameter
 combination runs. 
Table \ref{table3} shows the average MAE-POST and MAE-MED for
 all nine problem (averaged over the six hyperparameter combintations).
 

\begin{table}
\small
\begin{center}
\begin{tabular}{|c||c|c|c|c||c|c|c|c|}
\hline

 Problem & \multicolumn{4}{|c||}{MAE-POST} & \multicolumn{4}{|c|}{MAE-MED} \\
\hline\hline
	&	NEW	&	TITSIAS	& ESS & HMC &	NEW	&	TITSIAS	& ESS & HMC\\
Problem 1 & 0.1177 & 0.1176 & 0.1177 & 0.1153 & $1.4\times e^{-3}$	& $1.7\times e^{-3}$ & $2.6\times e^{-3}$ & $8.1\times e^{-3}$\\
Problem 2 &	0.0668 & 0.0676 & 0.06711 & 0.0673 & $3.2\times e^{-3}$	& $2.6\times e^{-3}$ & $3.1\times e^{-3}$ & $8.8\times e^{-3}$\\
Problem 3 &	0.0822	&	0.0830 & 0.0824 & 0.0837 & $2.3\times e^{-3}$	& $4.7\times e^{-3}$ & $3.1\times e^{-3}$ & $1.7\times e^{-2}$\\
Problem 4 &	0.1637 & 0.1641 & 0.1632 & 0.1685 & $1.6\times e^{-3}$	& $1.9\times e^{-3}$ & $3.3\times e^{-3}$ & $1.8\times e^{-2}$\\
Problem 5 &	0.2884 & 0.2883 & 0.2890 & 0.2872 & $1.5\times e^{-3}$	& $2.0\times e^{-3}$ & $3.0\times e^{-3}$ & $1.2\times e^{-2}$\\
Problem 6 &	0.0382 & 0.0384 & 0.0359 & 0.0367 & $7.2\times e^{-3}$	& $1.1\times e^{-2}$ & $4.1\times e^{-3}$ & $8.6\times e^{-3}$\\
Problem 7 &	-- & -- & -- & -- & $1.8\times e^{-3}$	& $2.1\times e^{-3}$ & $2.9\times e^{-3}$ & $1.3\times e^{-2}$	\\
Problem 8 &	-- & -- & -- & -- & $1.6\times e^{-3}$	& $1.9\times e^{-3}$ & $2.3\times e^{-3}$ & $9.7\times e^{-3}$	\\
Problem 9 &	-- & -- & -- & -- & $2.8\times e^{-3}$	& -- & $4.1\times e^{-3}$ & $2.2\times e^{-2}$	\\


\hline
\end{tabular} 
\end{center} 
\caption{The Mean Absolute Error 
between the Algorithms' Probability Estimates and the Bayesian
Posterior Probability (MAE-POST) and the Median of the Algorithms' Probabilities (MAE-MED)
over All Hyperparameter Sets. Note that for the Real-World Problems 7, 8, and 9 MAE-POST is not Available}  
\label{table3}
\end{table}
From the runs we note the following observations:
\begin{itemize}
\item The runs of all methods lead to 
close probability estimates for all the test patterns
and all the runs (typically an absolute error between the probability 
estimates and the median of probability estimates is of the order $10^{-3}$).
The exception is with HMC, which is a further from the other algorithms' estimates.
\item All methods lead to very similar mean absolute error with respect to
the Bayesian posterior error. This error is also fairly
low, especially if the size of the training set is large.
This is an indication that the source of the descrepancy
is probably the finite-sampled-ness of the training set,
rather than an inadequacy of the GPC model.
\item The TITSIAS method was considerably slower for
hyperparameter sets 4) and 6), and a little slower 
for hyperparameter sets 3) and 5). Even though we had fixed
the number of iterations for all hyperparameter sets (at 50,000),
it took about ten times as much to complete the run for sets 4) and 6) (compared to parameter sets 1) and 2)
and compared to the other methods).
It seems that lower values  of $\alpha$ lead to much slower runs for the TITSIAS method.
In contrast, the proposed Monte Carlo method yields similar speeds
for all parameter sets. 
Also, the TITSIAS method did not converge for Problem 9 (the USPS problem).
Even when reducing the number of samples to 10,000, it did not converge in more
than 24 hours of a run.
\item For most methods the runs take about a few minutes for a small 
problem (like $NTRAIN=NTEST=50$), and about ten minutes for a medium
problem (like $NTRAIN=NTEST=200$). 
Of course, this is with the exception
of the hyperparameter combinations that lead to slow runs for the TITSIAS method.
For example for Problem 3
($NTRAIN=NTEST=200$) the proposed method with $M=1,000,000$ 
took 12 minutes using Matlab on a computer featuring an Intel duo core I3 processor.
Again, it is hard to perform an accurate speed comparison between the algorithms, because
we do not know the ground truth.
\item The marginal likelihood of the proposed Monte Carlo algorithm 
and the ESS algorithm are very close. However, the HMC algorithm
produces a marginal likelihood that is somewhat different (often about 5 \% different, and for Problem 9 about 100 \%
different from that of the other two
algorithms). This indicates that
it fell short of fully converging. 
\end{itemize}

\subsection{Comments on the Results}
Working with general Bayesian methods often leads 
to high dimensional integrals.
Evaluating such integrals can sometimes be frustrating
because of the high dimensionality, and many Monte Carlo
approaches fail. The advantage of the proposed appraoch
is that it is very reliable. It basically works all
the time, as we have not encountered a failing run.
It is a fairly short algorithm, and is simple to code,
so this will cut down on development time.
The fact that it has no tuning parameter (other than
the number of Monte Carlo samples) also facilitates
applying the method and cuts down on time
consuming tuning runs.
The proposed algorithm has a main advantage compared
to the other MCMC-based algorithms. 
It is considerably faster for the problem 
of evaluating the marginal likelihood. This is important
because of its repeated evaluation during
the hyperparameter tuning step, 
and this makes it a very time-consuming process.
For this step it is sensible
to use a smaller number $M$ of Monte Carlo samples (for example 20,000 to 50,000). Exact evaluation
of the marginal likelihood will not much impact the optimization
outcome. It is evidenced by Table \ref{table2a} and Table \ref{table2b} that small size Monte Carlo
samples achieve very low error for the log marginal likelihood
(significantly lower than the error in the class posterior probabilities).
Also, running small samples could possibly yield three or four
best hyperparameter sets, for which on a closer look we rerun the method
using a larger $M$ to differentiate between them (a classic exploration versus exploitation problem).
Once the optimal parameters are obtained, in the classification
step we can then use a larger $M$ (for example 200,000 or 500,000),
because then, accuracy is important. 

An interesting observation is that the impact of the dimension of the integral $N$
on the accuracy of the proposed method is not that large. It seems that other factors weigh in more,
such as the structure of the covariance matrix, and the resulting conditional probabilities.
It is imperative to shuffle the class 1 and class 2 patterns, or interleave them
in a regular way, before constructing the covariance matrix. This will lead to well-behaved
conditional probabilities, and therefore better accuracy. 
  
\section{Conclusions}

In this paper we derived a new formulation that simplifies the multi-integral formula for
the Gaussian process classification (GPC) problem.
The formulation, given in terms of the ratio
of two multivariate Gaussian integrals,
gives new insights, and potentially opens
the door for better approximations.

We also developed a Monte Carlo method for the evaluation
of multivariate Gaussian integrals.
This allows us to obtain very close
to exact evaluation of the GPC probabilities
and the marginal likelihood function.
The proposed method is simple, reliable, and fast.
As such, it should be considered as a promising candidate for researchers 
to test, when attempting to obtain exact GPC probabilities.
\section*{Appendix}
The matrix $A_{22}$ can be written as
\begin{equation}
A_{22}=B+{{{\bf b}{\bf b}^T}\over{\sigma_*^2}}
\end{equation}
where
\begin{equation}
B=\pmatrix{1 & 0\cr
           0 & I+\Sigma^{-1}\cr
           }
\end{equation}
\begin{equation}
{\bf b}=\pmatrix{1\cr
           -{\bf a}\cr
           }
\end{equation}
Using the small rank adjustment matrix inversion lemma \cite{HORN85}, we get
\begin{equation}\label{A22MAT}
A_{22}^{-1}=B^{-1}-{{B^{-1}{\bf b}{\bf b}^T B^{-1}}\over{q}}
\end{equation}
where
\begin{equation}\label{Q-EQ}
q=\sigma_*^2+{\bf b}^TB^{-1}{\bf b}
\end{equation}
Substituting (\ref{A22MAT}) into Eq. (\ref{eq_A}), we get 
\begin{equation}\label{AMAT-20}
A=\pmatrix{0 & 0\cr  0 & I-C'(I+\Sigma^{-1})^{-1}C' \cr} +  {{\bf b'} {\bf b'}^T\over{q}}  
\end{equation}
where
\begin{eqnarray}
{\bf b'}=A_{12} B^{-1}{\bf b} &=&\pmatrix{-1 \cr +C'(I+\Sigma^{-1})^{-1}{\bf a} \cr}\\
&=& \pmatrix{-1 \cr +C'(I+\Sigma )^{-1} \Sigma_{X {\bf x}_*}\cr }
\end{eqnarray}
(The last equality follows from substituting for the variable ${\bf a}$ from Eq. (\ref{a-DEF}) (i.e. ${\bf a}={\Sigma^{-1}}\Sigma_{X {\bf x}_*}$),
and teleporting 
the resulting $\Sigma^{-1}$ into the bracketed expression $(I+\Sigma^{-1})^{-1}$.)
The variable $q$ can be simplified as follows:
\begin{eqnarray} 
q&=& \sigma_*^2 +  {\bf b}^T \pmatrix{1& 0 \cr 0 & (I+\Sigma^{-1})^{-1}} {\bf b}\\
&=& \sigma_*^2 +  1 + {\bf a}^T (I+\Sigma^{-1})^{-1} {\bf a} \\
&=& 1 + \Sigma_{{\bf x}_* {\bf x}_*}-\Sigma_{X {\bf x}_*}^T(I+\Sigma)^{-1} \Sigma_{X {\bf x}_*} \label{q-VAL}
\end{eqnarray}
The last equation follows from the definition of ${\bf a}$ (Eq. \ref{a-DEF}),
and the definition of  $\sigma_*^2$ Eq. (\ref{sigstar}), and several steps of simplification.

The first matrix in the RHS of Eq. (\ref{AMAT-20})
can be simplified further, by noting that
\begin{eqnarray}
I-C'(I+\Sigma^{-1})^{-1}C'&=& C'\Bigl[ I-(I+\Sigma^{-1})^{-1}\Bigl] C'\\
&=& C'(I+\Sigma^{-1})^{-1}\Bigl[ (I+\Sigma^{-1})-I\Bigl] C'\\
&=& C'(I+\Sigma)^{-1} C'
\end{eqnarray}
where we used the fact that $C'^2=I$ because it is a diagonal matrix of 1's and -1's.
We get the final formula for $A$, as follows:

\begin{equation}\label{A-FIN}
A=\pmatrix{ {1\over{q}} & \ \ \ & -{{\Sigma_{X {\bf x}_*}^T (I+\Sigma)^{-1} C'}\over{q}} \cr
-{{C'(I+\Sigma)^{-1} \Sigma_{X {\bf x}_*}}\over{q}} & \ \ \ &  C'(I+\Sigma)^{-1} C'+ {{C'(I+\Sigma)^{-1} \Sigma_{X {\bf x}_*} \Sigma_{X {\bf x}_*}^T (I+\Sigma)^{-1} C'}\over{q}}
\cr}
\end{equation}
Let us construct the following matrix, which we will subsequently try to invert.

\begin{equation}
R=\pmatrix{ {1+\Sigma_{{\bf x}_* {\bf x}_*}} & 
{\Sigma_{X {\bf x}_*}^T C'} \cr
{C' \Sigma_{X {\bf x}_*}} & 
C'(I+\Sigma )C' \cr }
\end{equation}
Using the partitioned matrix inverse theory \cite{HORN85},
we get
\begin{equation}\label{R-INV}
R^{-1}=\pmatrix{ {1\over{q}} & \ \ \ & -{{\Sigma_{X {\bf x}_*}^T (I+\Sigma)^{-1} C'}\over{q}} \cr
-{{C'(I+\Sigma)^{-1} \Sigma_{X {\bf x}_*}}\over{q}} & \ \ \ &  C'(I+\Sigma)^{-1} C'+ {{C'(I+\Sigma)^{-1} \Sigma_{X {\bf x}_*} \Sigma_{X {\bf x}_*}^T (I+\Sigma)^{-1} C'}\over{q}}
\cr}
\end{equation}
which is the same as $A$ in Eq. (\ref{A-FIN}), and that completes the proof.

\end{document}